\documentclass[10pt]{asme2ej}

\usepackage[T1]{fontenc}
\usepackage[utf8]{inputenc}
\usepackage{textcomp}
\usepackage{mathptmx}
\usepackage[numbers,sort,compress]{natbib}

\usepackage{graphicx}
\usepackage{overpic}
\usepackage[outline]{contour}
\contourlength{2pt}

\usepackage{algorithm}
\usepackage{algpseudocode}

\usepackage{enumitem}
\setlist{noitemsep,leftmargin=*}

\usepackage{amsmath,amsthm,amssymb}
\newtheorem{theorem}{Theorem}
\theoremstyle{definition}
\newtheorem{definition}{Definition}

\newtheorem*{remark*}{Remark}

\usepackage{hyperref}
\hypersetup{
  pdfauthor={Gábor Hegedüs, Josef Schicho, Hans-Peter Schröcker},%
  pdftitle={Four-Pose Synthesis of Angle-Symmetric 6R Linkages},%
  pdfkeywords={kinematic synthesis, 6R linkage, angle-symmetric
    linkage, motion polynomial, factorization},
  hidelinks,
}

\usepackage{todonotes}

\renewcommand{\DH}{\mathbb{DH}}
\newcommand{\SQ}{\mathcal{S}}
\newcommand{\SQF}{q}
\newcommand{\SE}[1][3]{\mathrm{SE}(#1)}
\newcommand{\RSet}{\mathbb{R}}
\newcommand{\eps}{\varepsilon}
\newcommand{\qi}{\mathbf{i}}
\newcommand{\qj}{\mathbf{j}}
\newcommand{\qk}{\mathbf{k}}
\newcommand{\peq}{\cong}

\newcommand{\abs}[1]{\vert #1 \vert}
\newcommand{\Norm}[1]{\Vert #1 \Vert}
\DeclareMathOperator{\arccot}{arccot}
\DeclareMathOperator{\lcoeff}{lc}

\DeclareMathOperator{\primal}{primal}

\newcommand{\bull}{\ensuremath{-}\hspace*{0.5em}}

\title{Four-Pose Synthesis of\\
  Angle-Symmetric 6R Linkages}

\makeatletter
\renewcommand{\author}[1]{%
    \global\advance\@authcnt1
    \expandafter\gdef\csname @auth\romannumeral\the\@authcnt\endcsname
    {\normalsize\sf\begin{tabular}[t]{@{}c@{}}\bfseries\sffamily#1\\[20pt]
     \end{tabular}}}
\makeatother

\author{Gábor Hegedüs
  \affiliation{
    Applied Mathematical Institute,\\
    Antal Bejczy Center for\\
    Intelligent Robotics, Obuda University\\
    hegedus.gabor@nik.uni-obuda.hu
  }
}
\author{Josef Schicho
  \affiliation{
    Johan Radon Institute for\\
    Computational and Applied Mathematics,\\
    Austrian Academy of Sciences\\
    josef.schicho@oeaw.ac.at
  }
}
\author{Hans-Peter Schröcker
  \affiliation{%
    Unit Geomtery and CAD,\\
    University Innsbruck,\\
    hans-peter.schroecker@uibk.ac.at
  }
}

\begin{document}

\maketitle

\begin{abstract}
  \itshape We use the recently introduced factorization theory of
  motion polynomials over the dual quaternions for the synthesis of
  closed kinematic loops with six revolute joints that visit four
  prescribed poses. Our approach admits either no or a one-parametric
  family of solutions. We suggest strategies for picking good
  solutions from this family.
\end{abstract}

\section{Introduction}

In \cite{brunnthaler05}, Brunnthaler et al. presented a method for
synthesizing closed kinematic chains of four revolute joints whose
coupler can visit three prescribed poses (position and orientation).
These linkages are also known under the name ``Bennett linkage''. By
that time, it was already known that three poses in general position
define a unique Bennett linkage. Other synthesis methods were
available \cite{veldkamp67,suh69,tsai73,perez04}. With hindsight, the
most important novelty of \cite{brunnthaler05} was the
characterization of the coupler motions of Bennett linkages as conics
in the Study quadric model of the direct Euclidean displacement group
$\SE$. Depending on the precise concept of ``Bennett linkages'', it
might be necessary to exclude certain conics, see \cite{hamann11}.
This interpretation allows to immediately construct the \emph{coupler
  motion} from the three given poses---even before the linkage itself
is determined.

In this article, we generalize the method of \cite{brunnthaler05} to
the four-pose synthesis of special kinematic loops of six revolute
joints (6R linkages). Our main tool is the recently developed
technique of factoring rational motions \cite{hegedus13}, or more
precisely, motion polynomials. This allows to decompose a rational
parametrized equation in Study parameters into the product of linear
factors. Each such factorization gives rise to an open kinematic
chain. Suitable combinations of open chains produce closed loop
linkages whose coupler follows the prescribed rational motion. If we
combine this factorization of rational motions with rational
interpolation techniques on a quadric, we obtain a framework for
mechanism synthesis. It requires the completion of three steps:
\begin{itemize}
\item \bull Interpolate $n$ given poses by a rational motion of degree $d$,
  parametrized by a motion polynomial $C(t)$.
\item \bull Factor the motion polynomial $C(t)$ and combine
  different factorizations to form a closed loop linkage.
\item \bull Pick a suitable linkage from the resulting family of solutions.
\end{itemize}
The mentioned synthesis of Bennett linkages corresponds to $n=3$ poses
and motion polynomials of degree $d=2$. The motion polynomial $C(t)$
admits two factorizations in two linear factors, each of which
determines one fixed and one moving axis. These axes can be determined
by geometric arguments as in \cite{brunnthaler05} or by a
straightforward algorithmic procedure (see
Algorithm~\ref{alg:factorization} below). The latter method extends to
motions of higher degree. An important aspect of the suggested
synthesis procedure is the fact that the mechanism is constructed from
its coupler motion. Thus, optimal synthesis is not restricted to
linkage characteristics only. It is possible and advisable to optimize
the coupler motion as well.

In this paper we are concerned with the case $n = 4$ and $d = 3$. The
reason for this is that, at least in general, higher degree motion
polynomials result in multi-looped spatial linkages which are probably
too complicated for most engineering applications. In case of $d = 3$
we obtain a certain class of overconstrained single looped 6R linkages
with a single degree of freedom.

We continue this text be recalling some basic concepts of motion
polynomials in \autoref{sec:dual-quaternions}. In
\autoref{sec:factorization} we describe algorithmic aspects of the
factorization algorithm of \cite{hegedus13}. In
\autoref{sec:interpolation} we show how to interpolate four points on
the Study quadric by twisted cubics and in \autoref{sec:synthesis} we
discuss some properties of the resulting linkages and make suggestions
for picking feasible solutions. \autoref{sec:example} illustrates the
complete algorithm at hand of a comprehensive example.

\section{Dual quaternions and motion polynomials}
\label{sec:dual-quaternions}

We parametrize the group $\SE$ of direct Euclidean displacements by
unit dual quaternions with non-vanishing primal part. The use of dual
quaternions in mechanism synthesis is not new. See \cite{perez03} for
an example and more pointers to the relevant literature. We assume
that the reader is familiar with the basic concepts and suggest
\cite{mccarthy11} as a further reference. The group $\SE$ is
isomorphic to the factor group of unit dual quaternions modulo
$\{\pm 1\}$. This means that every direct Euclidean displacement
corresponds to two unit dual quaternions $q$ and $-q$ of non-zero
primal part in such a way that composition of displacements
corresponds to dual quaternion multiplication.

In order to get rid of the ambiguity in the representation of elements
of $\SE$, one can factor by the real numbers and not just by $\{\pm
1\}$. In this way, we arrive at the real projective space $P^7$. The
unit norm condition reduces to the non-vanishing of the primal part
and the vanishing of the dual part. The latter, a homogeneous
quadratic form, defines the so-called \emph{Study quadric $\SQ$,} the
former defines the \emph{exceptional three-space $E$.} The image of
$\SE$ under this kinematic map is $\SQ \setminus E$. Homogeneous
coordinates in $P^7$ are called Study parameters. In this context, a
motion is a curve on $\SQ$, possibly with isolated points in $E$. We
are interested in rational motions. These are defined by polynomial
parametrizations in Study parameters or, equivalently, by motion
polynomials.

Essentially, a motion polynomial is the parametrized equation of a
rational motion in dual quaternions. More formally, we have:

\begin{definition}
  \label{def:1}
  A left polynomial $C$ with dual quaternion coefficients is called a
  \emph{motion polynomial} if $C\overline{C}$ is a real polynomial, if
  its leading coefficient equals $1$, and $\deg{C}$ is greater than
  zero.
\end{definition}

The condition on the leading coefficient is merely a technicality and
constitutes no loss of generality. We denote the set of all left
polynomials in the indeterminate $t$ and with dual quaternion
coefficients by $\DH[t]$. The non-commutativity of dual quaternion
multiplication entails some subtleties. To begin with, in the
multiplication of two polynomials in $\DH[t]$ the indeterminate $t$
commutes, \emph{by definition,} with all coefficients. This implies
that there is a difference between left polynomials, with coefficients
written to the left of the indeterminate, and right polynomials where
coefficients are on the right. We consistently use left polynomials
but often omit the additional word ``left''. The sequence of
polynomial multiplication matters and must strictly be obeyed in the
algorithms we present below.

A dual quaternion $h$ is called a rotation quaternion if it describes
a rotation and a translation quaternion, if it describes a
translation. Rotation quaternions are characterized by orthogonality
of primal and dual part (Study condition), vanishing dual scalar part
and non vanishing primal vector part. Translation quaternions are
characterized by the Study condition, vanishing dual scalar part,
vanishing primal vector part and non-vanishing primal scalar part. In
this text, we will usually not encounter translation quaternions as we
describe numeric procedures that produce translation quaternions with
negligible probability.

\section{Factorization of motion polynomials}
\label{sec:factorization}

A polynomial with real coefficients admits an essentially unique
factorization into linear factors over the complex numbers. The
situation for polynomials with dual quaternion coefficients is
fundamentally different. Because of the non-commutative
multiplication, a possible factorization is, in general, not unique.
The defining conditions for motion polynomials
(Definition~\ref{def:1}) guarantees the existence of several special
factorizations, at least in generic cases.

\begin{theorem}[\cite{hegedus13}]
  \label{th:1}
  For a generic motion polynomial $C \in \DH[t]$ of degree $n$ there
  exists $n!$ different factorizations $C(t) = (t-h_1)\cdots(t-h_n)$
  with rotation quaternions $h_1\ldots,h_n$.
\end{theorem}

Here, the term ``generic'' refers to the non-vanishing of the primal
part of $C(t)$. Usually, zeros of $\primal C$ turn some of the
rotation quaternions into translation quaternions. Motion polynomials
that admit no factorization do exist. Although their precise
characterization is not known, it is already clear that they are too
special to be of real relevance in synthesis procedures. The total
number $n!$ of factorizations comes from permutations of the conjugate
complex root pairs of the \emph{norm polynomial} $C\overline{C}$ of
degree $2n$ (compare Algorithm~\ref{alg:factorization} below). If some
of these root pairs coincide, the actual number of factorizations may
be less. Again, this non-generic case might be relevant to some
special synthesis problems but not in this article.

Any factorization $(t-h_1)\cdots(t-h_n)$ of a motion polynomial $C$
encodes an open $n$R-chain whose end-effector can follow the motion
parametrized by $C(t)$ with $t \in \RSet \cup \{\infty\}$. The
rotation quaternions $h_1,\ldots,h_n$ give its axes in the zero
position $C(\infty) = 1$. Different factorizations give rise to
different open chains that can be combined to form closed loop
linkages. Using factorization of motion polynomials, linkage synthesis
is turned into a rational interpolation problem on the Study quadric.
Rational interpolation on quadrics is a well-known topic in computer
aided design \cite{gfrerrer99,gfrerrer00}. Relevant for our purposes
are only low degree and low dimensional cases, whose geometry is
well-understood.

A constructive proof of \autoref{th:1} can be found in
\cite{hegedus13}. Here, we describe the corresponding factorization
algorithm. We denote the degree of $C \in \DH[t]$ by $\deg(C)$ and by
$\lcoeff(C)$ its leading coefficient. We use three auxiliary
procedures:
\begin{itemize}
\item \bull For a non-negative real polynomial $P$, $\Call{Factors}{P}$
  returns a list $[M_1,\ldots,M_n]$ of quadratic polynomials such that
  $P = M_1 \cdots M_n$ and each factor $M_i$ has at most a single real
  root. This list is unique up to permutation.
\item \bull For $A,B \in \DH[t]$ with $\lcoeff(B) = 1$, $Q =
  \Call{Quo}{A,B}$ and $R = \Call{Rem}{A,B}$ are the unique
  polynomials in $\DH[t]$ such that $A = QB + R$ (quotient and
  remainder of polynomial division). For their computation, a single
  procedure $\Call{QR}{A,B}$ can be used. It is shown in Algorithm~\ref{alg:qr}.
\end{itemize}
The function $\Call{Rmul}{D,P}$ right multiplies all left polynomials
in a list $D$ with the left polynomial $P$, the function
$\Call{Append}{R,D}$ appends a polynomial to a list of polynomials.

\begin{algorithm}[tb]
  \caption{Quotient and remainder of polynomial right division}
  \label{alg:qr}
  \begin{algorithmic}
    \Procedure{QR}{$A,B$}\Comment{Quotient and remainder of $A/B$;
      returns $Q$ and $R$ such that $A = QB + R$}
      \Require{$\lcoeff(B) = 1$}
      \State $Q \gets 0$
      \State $R \gets A$
      \While{$\deg(R) \ge \deg(B)$}\Comment{Ensure that $A = QB + R$ always holds true}
        \State $l \gets \lcoeff(R)$
        \State $Q \gets Q + l t^{\deg(R)-\deg(B)}$
        \State $R \gets R - B l t^{\deg(R)-\deg(B)}$
      \EndWhile
      \State\Return $Q,R$
    \EndProcedure
  \end{algorithmic}
\end{algorithm}

The computation of quotient and remainder works over arbitrary rings
and in both, exact or floating point arithmetic. This is important
because, in contrast to \cite{hegedus13}, we have to deal with
non-rational input data or with motion polynomials $C$ whose norm
polynomial $C\overline{C}$ has no factorization over the rational
numbers.

\begin{algorithm}[tb]
  \caption{Factorization of motion polynomials}\label{alg:factorization}
  \begin{algorithmic}
   \Procedure{Fac}{$C$} \Comment{All factorizations of $C$}
     \Require{$\lcoeff(C) = 1$, $C\overline{C} \in \RSet[t]$} \Comment{$C$ is monic motion polynomial}
     \State $n \gets \deg C$
     \If{$n = 0$}
       \State\Return{$C$}
     \EndIf
     \State $R \gets [\;]$\Comment{initialize empty list}
     \State $F \gets \Call{Factors}{C\overline{C}}$ \Comment{$F$ is list $[M_1,\ldots,M_n]$ of quadratic polynomials}
     \For{$i \gets 1,n$}
       \State $L \gets \Call{Rem}{C,M_i}$ \Comment{$L$ is linear left polynomial over $\DH$}
       \If{$\primal\lcoeff(L) = 0$}
         \State \textsc{Error!} \Comment{$\lcoeff(L)$ not invertible; factorization fails}
       \EndIf
       \State $h \gets -\lcoeff(L)^{-1}L(0)$ \Comment{$h$ is unique zero of $L$}
       \State $C' \gets \Call{Quo}{C,t-h}$ \Comment{remainder of polynomial division is zero}
       \State $D \gets [\Call{Fac}{C'}]$
       \State $D \gets \Call{Rmul}{D,t-h}$ \Comment{right multiply elements of $D$ with $t-h$}
       \State $R \gets \Call{Append}{R,D}$
     \EndFor
     \State \Return{$R$}
   \EndProcedure
  \end{algorithmic}
\end{algorithm}

Algorithm~\ref{alg:factorization} returns a list of all factorizations
of a given motion polynomial $C$. Our recursive implementation might
not be the most efficient. For example the call of
$\Call{Factors}{C\overline{C}}$ for each recursive call could be
avoided because it produces only a subset of the call at the previous
recursion level. Nonetheless, Algorithm~\ref{alg:factorization} is
easy to implement and fast for reasonably small polynomial degrees.
The pseudocode of Algorithm~\ref{alg:factorization} shows that every
factor $(t-h_i)$ in a particular factorization of $C$ can be
associated with a quadratic factor $M_i$ of $C\overline{C}$. The
complete factorization of $C$ corresponds to a permutation of the
factors of $C\overline{C}$. Moreover, we see that the factorization
fails if the linear remainder polynomial has a non-invertible leading
coefficient. In \cite{hegedus13} we proved that this never happens in
generic cases. We will ignore this possibility in the present
investigation.

\section{Cubic interpolation on the Study quadric}
\label{sec:interpolation}

In \autoref{sec:factorization} we saw how to factor a motion
polynomial. Here, we discuss the construction of such polynomials to
four given points $p_1,p_2,p_3,p_4$ on the Study quadric $\SQ$. More
precisely, we want to find a cubic motion polynomial $C(t)$ such that
there exist values $t_1,t_2,t_3,t_4 \in \RSet \cup \{\infty\}$ with
$C(t_i) \peq p_i$ for $i \in \{1,2,3,4\}$. Here, the symbol ``$\peq$''
means equality in projective sense, that is, up to multiplication with
a non-zero real scalar.

Since the Study parameters are homogeneous, this can be seen as a
particular instance of rational interpolation problem on quadrics
\cite{gfrerrer99,gfrerrer00}. We are not
interested in all aspect of this topic. In particular, the restriction
to interpolants of degree $n = 3$ is a considerable simplification.
The case $n = 2$ (interpolation by conics) is even simpler and there
is no need to address it here. It has been extensively treated in
\cite{brunnthaler05,schroecker09,hamann11}.

We will study in more detail the interpolation of four points $p_1,
p_2, p_3, p_4$ on the Study quadric $\SQ \subset P^7$ by a rational
cubic interpolant $C(t)$ contained in $\SQ$. Doing so, we will always
assume that the four points $p_1, p_2, p_3, p_4$ have a projective
span $P$ of dimension three. In order to exclude spherical or planar
motions and particularities of translational joints, we assume that
$P$ is not contained in $\SQ$ and does not intersect the exceptional
generator $E$. This last assumption is not really necessary but
simplifies our exposition. Finally, it is helpful to set $p_1 = 1$.
This entails $t_1 = \infty$ and can always be achieved by a suitable
choice of coordinates.

Several methods for solving the interpolation problem are conceivable.
We describe an approach that fits well to our dual quaternion based
setting.

\subsection{Existence of solutions}

It is possible that no solution exists. This happens if the
intersection $Q = P \cap \SQ$ contains no real twisted cubics. Since
$Q$ is a real quadric in a projective three space, a simple criterion
for the existence of cubics on $Q$ is the existence of straight lines
on $Q$. We can test this by computing the straight lines on $Q$
through $p_1 = 1$. Any point $k$ on a straight line $L \subset \SQ$
through $1$ is generically a rotation quaternion or, in special cases,
a translation quaternionk In our description we ignore the latter
possibility. Rotation quaternions have a purely vectorial dual part.
Different points on the same line $L$ correspond to different
rotations about the same axes. The involutory rotations (half-turns)
are characterized by a purely vectorial primal part.

In order to compute the half-turn quaternions in $P$, we set $k =
x_1p_1 + x_2p_2 + x_3p_3 + x_4p_4$. The two half-turn conditions
require the vanishing of the scalar part of $k$. This gives two linear
conditions, the Study quadric condition gives one quadratic
conditions. Thus, we end up with a system of a quadratic and two
linear equations for the homogeneous unknowns $x_1, x_2, x_3, x_4$. It
is easy to compute its up to two real solutions. In case of no real
solution, the interpolation problem cannot be solved. The case of two
real solutions is dealt with in the next section. The non-generic case
of one solution is amenable to synthesis in the line of this article
but, for reasons of compact presentation, will be ignored in the
following.

\subsection{Computing all solutions}
\label{sec:computing-solutions}

We did not only compute the two half-turns $k_1,k_2 \in Q = P \cap S$
in order to test existence of solutions. We can also use them for
actually computing interpolants. The quadric $Q$ is doubly ruled, the
span $K_1$ of $1$ and $k_1$ belongs to one family of rulings and the
span $K_2$ of $1$ and $k_2$ to the other. Our aim is to find a twisted
cubic interpolating $p_1=1,p_2,p_3,p_4$. It is well-known that each
such cubic intersects every member of one family of rulings in one
point and each member of the other family in two points
\cite[p.~301]{salmon82}. Accordingly, we have to deal with two
completely symmetric situations. We describe the case where the cubic
intersects $K_1$ twice and $K_2$ once.

The second intersection point $p_5$ of $K_1$ and the cubic can be
written as $p_5 = \lambda - k_1$ with $\lambda \in \RSet \cup
\{\infty\}$. We use $\lambda$ to parametrize all solution cubics in
this family. The key observation is now that any rational cubic
parametrization of the sought curve is induced, via the rulings in the
family of $K_2$, by a linear rational parametrization of any ruling in
the family of $K_1$. Hence, we compute parameter values $t_i$ such
that $p'_i \peq t_i - k_1$ is the intersection point of a ruling
through $p_i$ with $K_1$ for $i \in \{2,3,4\}$. The values
$t_2,t_3,t_4$ will be the parameter values at which the cubic curve
interpolates the given points on the Study quadric. This is
necessarily so, up to admissible change of parameter via a fractional
linear re-parametrization. For the computation of $t_i$ we have to
evaluate the linear condition $\SQF(t_i-k_1, p_i) = 0$ where $\SQF$ is
the quadratic form associated to the Study quadric. The value $t_i =
\infty$ cannot occur because of $p_1 \neq p_i$, $i \in \{2,3,4\}$.
Setting $t_1 = \infty$, we end up with five points
$p_1=1,p_2,p_3,p_4,p_5$ and corresponding parameter values
$t_1=\infty,t_2,t_3,t_4,\lambda$, chosen in such a way that a cubic
interpolant exists. Now we can use a standard interpolation algorithm
to find a cubic interpolant $C$ such that $C(t_i) \peq p_i$ for $i \in
\{1,\ldots,4\}$ and $C(\lambda) \peq p_5$. This interpolant still
depends on the parameter $\lambda$. By construction, the polynomial
$C$ is a motion polynomial.


\section{Synthesis}
\label{sec:synthesis}

By combining the results of \autoref{sec:factorization} and
\ref{sec:computing-solutions} we are, under certain mild assumptions,
able to interpolate four given points $p_1=1,p_2,p_3,p_4 \in \SQ$ by
two one-parametric families of rational cubic curves, parametrized by
motion polynomials $C_\lambda(t)$ and $D_\mu(t)$, and factor each of
these cubic polynomials, in six different ways, as
$(t-h_1)(t-h_2)(t-h_3)$ with rotation quaternions $h_1,h_2,h_3$. Each
triple $(h_1,h_2,h_3)$ of rotation quaternions corresponds to an open
3R chain that can reach the given poses $p_1,p_2,p_3,p_4$. Suitably
combining two such chains gives a candidate 6R linkage for our
synthesis problem.

However, there are invalid combinations of open 3R chains that should
be avoided. Recall that each factorization obtained by
Algorithm~\ref{alg:factorization} corresponds to an ordering of the
set $\{M_1,M_2,M_3\}$ of quadratic factors of the norm polynomial
$C\overline{C}$. We must not combine factorizations with the same
quadratic factor at the first or the last position, as this produces a
dangling link attached to either the end-effector or the base. On the
level of factorizations, this can be distinguished by identical
factors on the right or on the left. Six among the 15 possible
combinations are afflicted with this defect. In order to exclude
invalid linkages a priori, it is necessary that the
$\Call{Factors}{C\overline{C}}$ procedure returns a list of factors
that is sorted with respect to some total ordering relation. Thus, a
consistent ordering of factors in recursive calls of $\Call{Fac}{C}$
is guaranteed. In this case, Algorithm~\ref{alg:factorization} returns
a list $R = [R_1,\ldots,R_6]$ of six factorizations in which the index
pairs $(1,4)$, $(1,5)$, $(1,6)$, $(2,3)$, $(2,4)$, $(2,6)$, $(3,5)$,
$(3,6)$, and $(4,5)$ represent valid combinations. Three of these
pairs, $(1,6)$, $(2,4)$, and $(3,5)$, give rise to ``general angle
symmetric linkages'' and six give rise to linkages of type ``Waldron's
double Bennett hybrid'' \cite{hegedus13,li13}.

At this point, we are left with two one-parametric families of six
factorizations, each giving rise to nine candidate linkage. We should
thus provide some method to pick ``good'' or at least ``suitable''
linkages from this variety. Before embarking on this, it must be said
that a mathematical optimization on either of the families of
solutions is problematic. In the description of
\autoref{sec:computing-solutions}, one family is parametrized by
$\lambda \in \RSet \cup \{\infty\}$. Since the actual computation of a
solution requires the factorization of a real polynomial of degree six
and a subsequent polynomial division, differentiation with respect to
$\lambda$ is difficult if not infeasible. However, since the solution
space is only of dimension one, a straightforward line search is
possible with reasonable effort.

\subsection{Interpolation order}
\label{sec:interpolation-order}

An important aspect in many synthesis problems is to guarantee a
certain interpolation sequence. In the instance we describe, this is,
unfortunately, not possible. From the construction in
\autoref{sec:computing-solutions}, it is obvious that the
interpolation times $t_1 = \infty, t_2, t_3, t_4$ for the four given
points are independent from the parameter $\lambda$ that determines
the cubic interpolant. Within one family of solutions, the given poses
are visited at exactly the same parameter times. The free parameters
alone cannot be used to resolve a possible order defect. The only way
to remedy this, is to change the input poses $p_1,p_2,p_3,p_4$.

\subsection{Coupler motion}
\label{sec:coupler-motion}

A specialty of the proposed procedure is the intermediate construction
of possible end-effector motions from which the mechanism is
synthesized. This adds an important ingredient to the problem of
finding optimal linkages. Not only should certain linkage
characteristics be satisfactory but the end-effector motion should be
acceptable as well. This brings our topic in the vicinity of motion
fairing \cite{fang98,hsieh03} and the construction of fair motions
\cite{hofer04}. However, applicability of these established research
topics to our problem should not be over-estimated. We do not
recommend to waste our single free parameter on a fair coupler motion
alone. It should also be used to produce satisfactory linkage
properties. Thus, we use a suitable fairness functional only for
eliminating linkages with bad coupler motion. The example in
\autoref{sec:example} below shows how this might work in applications.
It is for this reason, that particular features or possible defects of
the fairness functional, like dependence on chosen coordinate frames
or disparity of translational and rotational units, are of minor
relevance to us. In our example in \autoref{sec:example} we use a very
simple energy functional on the trajectories of certain feature
points.

\subsection{Linkage properties}
\label{sec:linkage-properties}

Once candidate linkages with unfavorable end-effector motion have been
eliminated, we can turn our attention to properties of the linkage
itself. We suggest to compute the usual Denavit-Hartenberg
parameters distances, angles, and offsets. From distances and
offsets, it is possible to estimate the overall size of the linkage
and exclude excessively large solutions. Small angles indicate a
relatively ``planar'' spatial linkage, which might also be desirable
in some cases.

While these are general considerations that also pertain to other
types of linkages, we also suggest a quality measure that is tailored
to linkages synthesized by factorization of cubic motion polynomials.
It is related to the joint angles $\varphi_1 = 0$ and $\varphi_4$ at
which the first and the last input pose are attained. In order to
understand their relation to the interpolation parameter times $t_1 =
\infty$ and $t_4$, we consider a rotation quaternion $h$ and the
parametrization $t-h$ of its rotations. The rotation angle $\varphi$
at parameter time $t \in \RSet \cup \{\infty\}$ is given by the
formula
\begin{equation}
  \label{eq:1}
  \varphi = 2\arccot \frac{t - r}{\sqrt{4s-r^2}} 
  \quad \text{where} \quad
  r = -h - \overline{h},\ s = h\overline{h}.
\end{equation}
This formula has interesting consequences. The polynomial $M = t^2
+rt + s$ is the \emph{minimal polynomial} of $h$---the unique
monic left polynomial of degree two with $M(h) = 0$. In
\cite{hegedus13}, we showed that if $t-h$ occurs in the factorization
of a motion polynomial $C$, than $M$ is the corresponding factor of
$C\overline{C}$. By Algorithm~\ref{alg:factorization}, the rotation
angles in two open 3R chains obtained through factorization of the
same motion polynomial are permutations of each other. This implies
existence of three joint pairs whose \emph{input-output relation is
  the identity.} This was also observed in \cite{li13} and is the
reason for the title of this paper.

Equation~\eqref{eq:1} implies that the rotation angle is strictly
monotone and the linkage has full cycle mobility, at least
generically. (These properties cannot be guaranteed if two consecutive
revolute axes coincide. The rotation angle in this joint is then the
sum of two rotation angles which may result in rocking joints.) This
behavior of the joint angle functions is certainly a pleasant property
of the synthesized linkages. It also suggests to use the angle
increment in each joint between the first and last pose as a possible
optimization target. In other words, the motion from $p_1$ to $p_4$
via $p_2$ and $p_3$ should be accomplished by small rotations in the
six joints. Of course, this only makes sense, if the resulting linkage
is not required to perform the full cycle end-effector motion. Since,
by construction, the first pose is the identity and corresponds to
zero rotation in all joints, we have to minimize the norm of the
vector $(\omega_1,\omega_2,\omega_3)$ of joint angles associated with
$p_4 = C(s_4)$. For this approach the following remarks hold true:
\begin{itemize}
\item \bull The set of rotation angles is independent from the
  particular factorization of $C$. In fact, it can be computed from
  the factors of $C\overline{C}$ alone and without running
  Algorithm~\ref{alg:factorization}.
\item \bull When using the maximum norm, the angle $\omega_i$ can
  be replaced by
  \begin{equation}
    \label{eq:2}
    m(h_i) = \frac{\abs{r}}{\sqrt{4s-r^2}}
  \end{equation}
  where $r = -h_i - \overline{h_i}$, $s = h_i\overline{h_i}$, and
  $h_i$ is the corresponding rotation quaternion. Again, $m(h_i)$ can
  be computed directly from the factors of the norm
  polynomial~$C\overline{C}$. We call it the factor's \emph{angle
    characteristic}. The assumption is that small rotations and
  sensible axis layout also produce good coupler motions in the range
  between $p_1$ and~$p_4$.
\item \bull An optimal solution in this sense is associated to the
  motion polynomial $C$, not to a linkage. It is still necessary to
  make a choice from the nine linkages produced by the different
  factorizations of~$C$.
\end{itemize}

\section{An example}
\label{sec:example}

In this section we discuss different aspects of a particular synthesis
problem. It is meant as an illustration of the general synthesis
process and also as an example of how to incorporate the ideas of the
preceding section in a synthesis pipeline. We want to synthesize an
overconstrained 6R linkage to the four poses
\begin{equation*}
  \begin{aligned}
    p_1 &= 1,\\
    p_2 &= 
-0.575+0.374\eps+(0.598-0.194\eps)\qi+(0.397+0.310\eps)\qj+(0.393+0.529\eps)\qk
,\\
    p_3 &= 
-0.312+0.939\eps+(0.903+0.116\eps)\qi+(0.189+0.219\eps)\qj+(0.225+0.653\eps)\qk
,\\
    p_4 &= 
-0.688+0.808\eps+(0.719+0.678\eps)\qi-(0.098+0.686\eps)\qj+(0.012+0.086\eps)\qk
.
  \end{aligned}
\end{equation*}
At first, we compute the two half-turn quaternions $k_1,$ $k_2$ in the
span of $p_1$, $p_2$, $p_3$, and $p_4$. They are
\begin{equation*}
  \begin{aligned}
    k_1 &= 
(0.546-0.115\eps)\qi+(0.583-0.174\eps)\qj+(0.602+0.273\eps)\qk
,\\
    k_2 &= 
(0.810+1.575\eps)\qi+(0.252-3.011\eps)\qj+(0.530-0.973\eps)\qk
.
  \end{aligned}
\end{equation*}
Their existence already implies that the interpolation problem admits
solutions.

Now we have to construct interpolating cubic motion polynomials to the
given poses. The theory predicts two one parametric families. The
first step is the computation of possible parameter quadruples
$(u_1,u_2,u_3,u_4)$ and $(v_1,v_2,v_3,v_4)$ for each family. Denote by
$\SQF$ the quadratic form associated to the Study quadric. The
parameter values are computed from the conditions $\SQF(p_i, u_i-k_1)
= \SQF(p_i, v_i-k_2) = 0$ for $ i \in \{2,3,4\}$. For each input pose,
this gives a system of seven linear equations with exactly one
solution. In our example, these values are
\begin{alignat*}{4}
  u_1 &= \infty,\quad&
  u_2 &= 
0.660
,\quad&
  u_3 &= 
0.368
,\quad&
  u_4 &= 
-0.034
;\\
  v_1 &= \infty,\quad&
  v_2 &= 
-0.294
,\quad&
  v_3 &= 
0.304
,\quad&
  v_4 &= 
0.575
.
\end{alignat*}
We added the parameter values $u_1$ and $v_1$ which, by construction,
are infinity. The parameter space for one-parametric rational motions
is the projective line. Hence, both quadruples visit the given poses
in the order implied by their numbering. \emph{Neither of the two
  families suffers from an order defect.}

Now we want to find rational cubics $C(u)$ and $D(v)$ such that
$C(u_i) \peq D(v_i) \peq p_i$ for $i \in \{1,2,3,4\}$. We only
describe the computation of $C(u)$. In order to avoid dealing with the
parameter value $u_1 = \infty$ we re-parametrize according to $s(u) =
u^{-1}$ and compute the cubic Lagrange interpolant $C(s)$ to the
points $w_ip_i$, $i \in \{1,2,3,4\}$. The weights $w_1,w_2,w_3,w_4$
still have to be determined. For that purpose, we introduce a fifth
point $p_5 = \lambda - h_1$ that depends on a free parameter
$\lambda$. The condition $C(\lambda^{-1}) \peq p_5$ gives us eight
linear equations for $w_1,w_2,w_3,w_4$. By construction, they have a
unique solution that still depends on $\lambda$. In other words, we
have found a one-parameter family of interpolating cubics
$C_\lambda(s)$. We re-substitute $s = u^{-1}$ and multiply away the
denominator $u^3$ to arrive at the input $C_\lambda(u)$ for the next
step. Similarly, we obtain a second family $D_\mu(v)$ of solutions.
Their formulas are too long to be displayed.

\begin{remark*}
  It is possible to extend the synthesis to geometric Hermite
  interpolation where we do not interpolate poses, but pose plus
  instantaneous velocity or even pose plus instantaneous velocity plus
  acceleration. The only necessary adaption is the replacement of
  Lagrange interpolation by Hermite interpolation. The calculations do
  not change much.
\end{remark*}

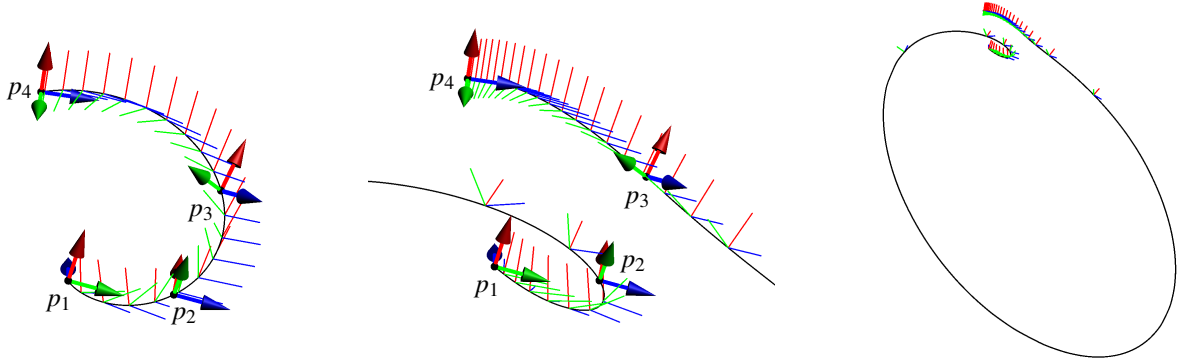
\begin{figure}
  \centering
  \begin{minipage}{0.30\linewidth}
    \centering
    \begin{overpic}{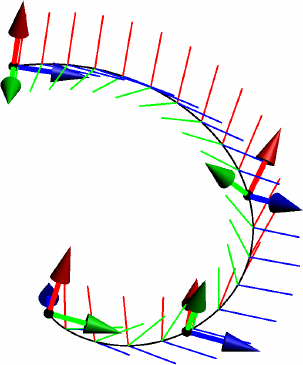}
      \put(5,5){$p_1$}
      \put(50,0){\contour{white}{$p_2$}}
      \put(56,37){\contour{white}{$p_3$}}
      \put(-9,81){$p_4$}
    \end{overpic}
  \end{minipage}%
  \begin{minipage}{0.35\linewidth}
    \centering
    \begin{overpic}[trim=150 555 245 0,clip]{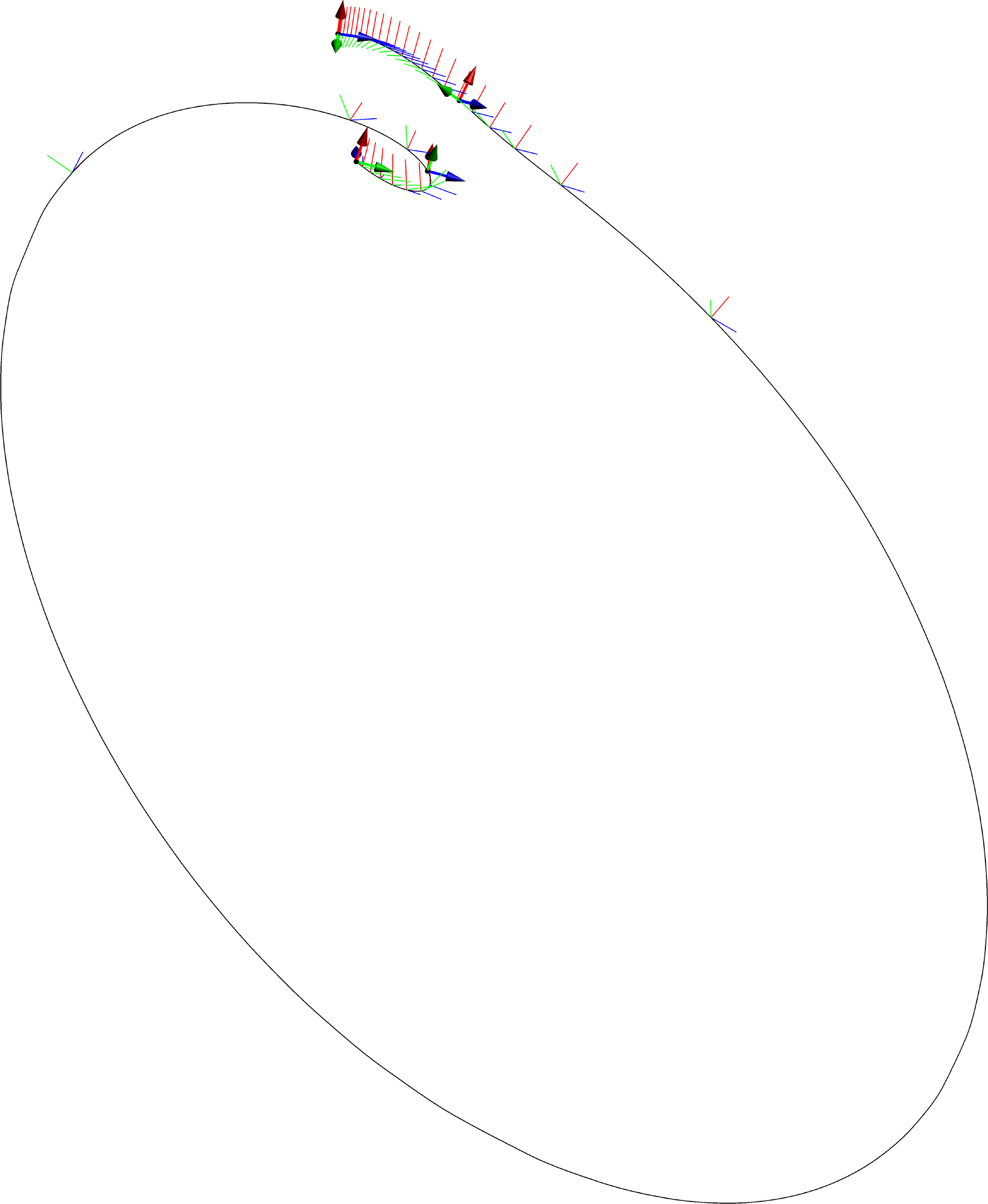}
      \put(26,9){$p_1$}
      \put(62,15){$p_2$}
      \put(63,31){$p_3$}
      \put(15,60){$p_4$}
    \end{overpic}
  \end{minipage}%
  \begin{minipage}{0.35\linewidth}
    \centering
    \includegraphics{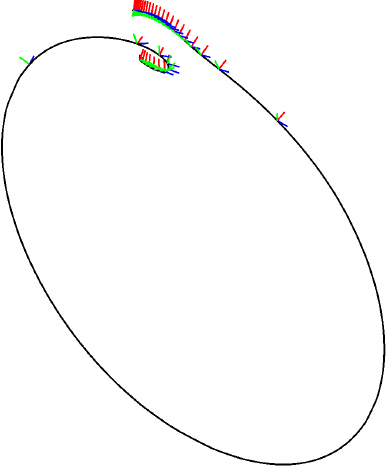}
  \end{minipage}
  \caption{Valid end-effector motions of different fairness}
  \label{fig:fairness}
\end{figure}

Let us now illustrate the importance of filtering out motions of poor
fairness. \autoref{fig:fairness} displays a fair candidate motion on
the left, a motion with bad fairness in the middle and the same bad
motion in a different scaling on the right. The poses to be
interpolated are indicated as well. It is obvious that the first
motion is preferable. The fairness criterion we use in this example is
\begin{equation}
  \label{eq:3}
  F(\lambda) = \sum_{i=0}^3 \int_{t_0}^{t_1} \Norm{x'_i(t)} \;\mathrm{d}t
\end{equation}
where $x_0(t)$, $x_1(t)$, $x_2(t)$ and $x_3(t)$ are the trajectories
of origin and the unit points on the three coordinate axes with
respect to the motion defined by $C_\lambda$. In other words, we
measure fairness as sum of arc-lengths of certain trajectories. The
trajectory of the origin is displayed in \autoref{fig:fairness}. The
arc-length as a measure of fairness is not very popular because of its
tendency to straighten curve segments. In our case, it seems to be an
acceptable choice because the single free parameter $\lambda$ is not
enough to ``over-optimization'' towards a kinky appearance.

We do not pick the optimal solution of \autoref{fig:fairness}.
Instead, we use \eqref{eq:3} to eliminate solutions of bad fairness.
In order to do so, we compute the fairness measures for a reasonable
discretization of the solution space and only accept good solutions.
In our example, we considered solutions in the first quartile as
acceptable. This gives the two ranges
$[
27.609
, 
28.629
]$ and
$[
26.917
, 
72.816
]$
for the fairness of motions in the discretized sets
$\{C_{\lambda_0},\ldots,C_{\lambda_m}\}$ and
$\{D_{\mu_0},\ldots,D_{\mu_n}\}$, respectively. Since the optimal
motion quality is comparable in both sets, we subject both to further
scrutiny but restrict our description to the first set.

For each of the remaining motion polynomials $C_{\lambda_i}$, we
compute quadratic irreducible polynomials $M_{i,1}$, $M_{i,2}$, $M_{i,3}
\in \RSet[t]$ such that the norm polynomial
$C_{\lambda_i}\overline{C_{\lambda_i}}$ equals $M_{i,1}M_{i,2}M_{i,3}$.
To each quadratic factor $M_{i,j} = t^2+r_{i,j}t+s_{i,j}$ we associate
its angle characteristic, given by \eqref{eq:2}. It is our aim to keep
the total rotation in the linkage small. For that purpose we pick the
curve $C_\lambda$ and $D_\mu$ as respective minimizers of their
maximal angle characteristics:
\begin{gather*}
  C(u) = 
0.050-0.065\eps-(0.055+0.052\eps)\qi+(0.010+0.048\eps)\qj+(0.002-0.008\eps)\qk+\\(-0.104-0.063\eps+(0.035+0.045\eps)\qi+(0.064-0.160\eps)\qj+(0.075-0.054\eps)\qk)t+\\(-0.242-0.000\eps-(0.321-0.177\eps)\qi-(0.381+0.071\eps)\qj-(0.377+0.247\eps)\qk)t^2+t^3
,\\
  D(v) = 
0.260-0.336\eps-(0.411-0.112\eps)\qi-(0.208+0.264\eps)\qj-(0.205+0.384\eps)\qk+\\(-0.884+0.115\eps+(0.511-0.380\eps)\qi+(0.559+0.389\eps)\qj+(0.533+0.525\eps)\qk)t+\\(0.870-0.000\eps-(0.374+0.363\eps)\qi-(0.244-0.815\eps)\qj-(0.320-0.162\eps)\qk)t^2+t^3
.
\end{gather*}
The fairness values are $F(\lambda) = 
28.629
$ and
$F(\mu) = 
36.298
$, respectively; the maximal angle
characteristics are $
1.268
$ and
$
2.041
$, respectively. The solution $C(u)$ is
preferable in both respects. Hence, we discard $D(v)$.

\begin{figure}
  \begin{minipage}[t]{0.5\linewidth}
    \rule{\linewidth}{0pt}
    \centering
    \includegraphics{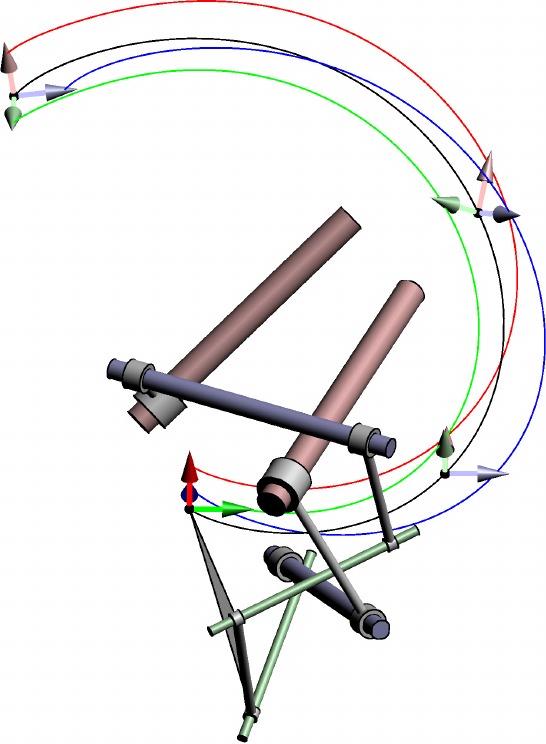}
  \end{minipage}%
  \begin{minipage}[t]{0.5\linewidth}
    \rule{\linewidth}{0pt}
    \centering
    \includegraphics{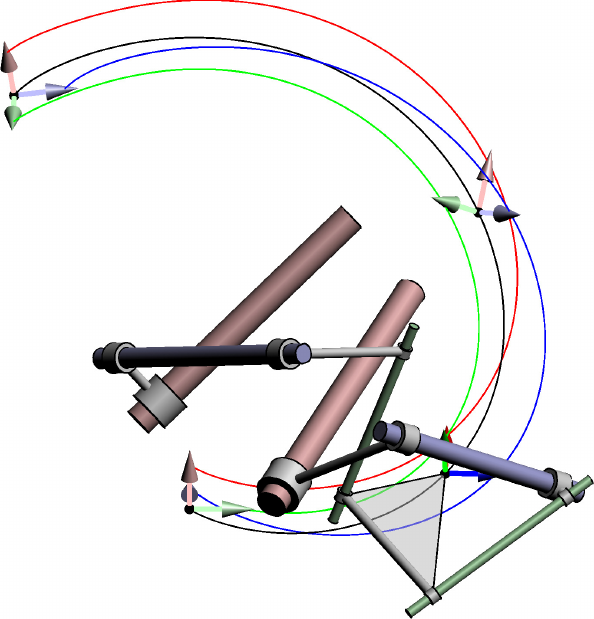}
  \end{minipage}\\[1ex]
  \begin{minipage}[t]{0.5\linewidth}
    \rule{\linewidth}{0pt}
    \centering
    \includegraphics{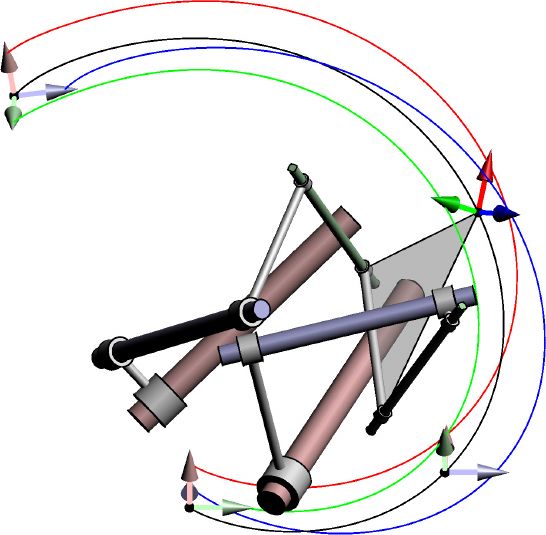}
  \end{minipage}%
  \begin{minipage}[t]{0.5\linewidth}
    \rule{\linewidth}{0pt}
    \centering
    \includegraphics{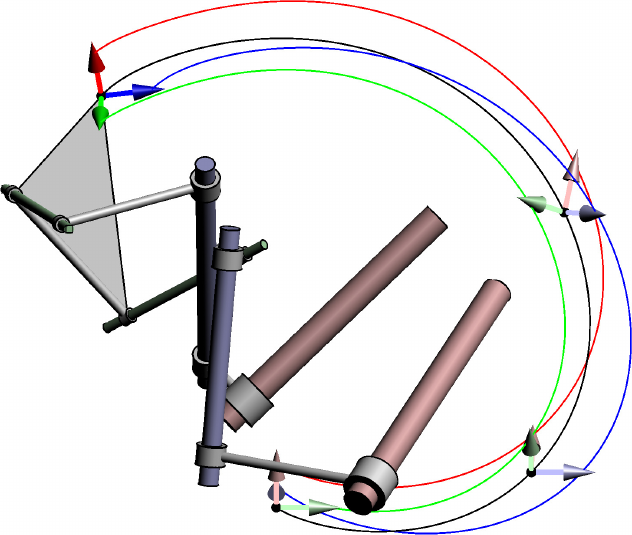}
  \end{minipage}
  \caption{The solution to our synthesis problem}
  \label{fig:solution}
\end{figure}

Finally, we have to decide upon a suitable linkage among the nine
linkages with coupler motion $C(u)$. We do this by minimizing the sum
of distances in the linkage. The solution linkage and the motion of
the end-effector coordinate system are visualized in
\autoref{fig:solution}.

\section{Synopsis}

We presented a synthesis procedure for a certain type of
overconstrained spatial 6R linkages that can visit four given poses.
To the best of our knowledge, this is the first reasonable general and
flexible synthesis method for 6R linkages. In lack of a concrete
application, some aspects of our description necessarily remained
rather general. We trust, however, that the example in
\autoref{sec:example} contains sufficient information and ideas to
apply the suggested procedure to engineering tasks.

Our scheme obviously extend to more input poses and higher degree
motion polynomials, however at the cost of increasing engineering
complexity. Future research will focus on factorization of higher
degree coupler motions with coinciding consecutive axes, thus keeping
small the number of joints while allowing for more input poses.
Another line of research is the synthesis of linkages whose coupler
motion satisfies certain incidence constraints like point on line or
point on sphere. This amounts to finding cubic curves on constraint
varieties in the Study quadric.

\section*{Acknowledgments}

This work was supported by the Austrian Science Fund (FWF): P~23831-N13.



\end{document}